\documentclass[times, review, 10pt]{elsarticle}

\usepackage{amsmath,amsfonts,amssymb}
\usepackage{algorithm}
\usepackage{algpseudocode}
\usepackage{array}
\usepackage{booktabs}
\usepackage{float}
\usepackage[section]{placeins}
\usepackage{graphicx}
\usepackage{textcomp}
\usepackage{verbatim}
\usepackage{xcolor}
\usepackage{soul}
\usepackage[font=footnotesize]{caption}
\usepackage[font=footnotesize]{subfig}
\usepackage{etoolbox}
\usepackage{multirow}

\journal{Pattern Recognition}

\makeatletter
\def\ps@pprintTitle{%
  \let\@oddhead\@empty
  \let\@evenhead\@empty
  \def\@oddfoot{%
    \footnotesize\itshape
    Preprint submitted to
    \ifx\@journal\@empty
      Elsevier
    \else
      \@journal
    \fi
    \hfill\today
  }%
  \let\@evenfoot\@oddfoot
}
\makeatother

\sethlcolor{yellow}

\AtBeginEnvironment{table}{\normalsize}
\AtBeginEnvironment{tabular}{\normalsize}
\AtBeginEnvironment{algorithm}{\normalsize}


\begin{document}

\begin{frontmatter}

\title{DAC-Pose: Dual-Agent Collaborative Framework for Pose-guided Human Generation\tnoteref{t1}}

\author[aff1,aff2,aff3]{Haotian Yang}
\ead{yang.haotian@foxmail.com}

\author[aff2,aff3]{Zhile Yang}
\ead{zl.yang@siat.ac.cn}

\author[aff4]{Huiyu Zhou}
\ead{hz143@leicester.ac.uk}

\author[aff1]{Xin Sun\corref{cor1}}
\ead{sunxin1984@ieee.org}
\cortext[cor1]{Corresponding author.}

\affiliation[aff1]{organization={Faculty of Data Science, City University of Macau},
  city={Macao},
  country={China}}

\affiliation[aff2]{organization={Shenzhen University of Advanced Technology},
  city={Shenzhen},
  country={China}}

\affiliation[aff3]{organization={Shenzhen Institute of Advanced Technology, Chinese Academy of Sciences},
  city={Shenzhen},
  country={China}}

\affiliation[aff4]{
  organization={School of Computing and Mathematical Sciences, University of Leicester},
  city={Leicester},
  country={United Kingdom}
}



\begin{abstract}AI agents have emerged as a powerful new paradigm in generative image synthesis, enabling systems to perform complex semantic reasoning rather than passive pixel-level mapping. In pose-guided human generation, conventional methods inevitably produce severe visual artifacts under drastic viewpoint shifts, fundamentally because they lack the cognitive capacity to logically deduce unseen regions and model complex spatial deformations. To bridge this gap, we propose DAC-Pose, a novel agent-driven multimodal framework that reformulates single-view human generation as a collaborative dual-agent system. DAC-Pose integrates two complementary components, i.e., Prior Semantic Reasoning (PSR) agent and Discrepancy-Aware Visual Encoding (DAVE) agent. Functioning as a cognitive engine, the PSR utilizes collaborative reasoning to deduce the fine-grained attributes of unseen regions. Concurrently, acting as a specialized visual perception agent, DAVE quantifies and encodes viewpoint-induced spatial misalignments, continuously feeding robust spatial constraints back into the generative process. This autonomous feedback loop between semantic deduction and visual perception ensures high-fidelity detail synthesis. Extensive experiments on the DeepFashion and Market-1501 benchmarks validate the superiority of our agent-driven paradigm. Notably, DAC-Pose excels in preserving unprecedented texture alignment and identity consistency under drastic viewpoint changes. The code is available at https://github.com/AIVRC/DAC-Pose.git.
\end{abstract}





\end{frontmatter}

\section{Introduction}

The integration of autonomous AI agents transforms image synthesis from passive pixel-level mapping to active semantic reasoning. A compelling and highly challenging application of this paradigm is pose-guided human image generation, which requires synthesizing photorealistic individuals in a target pose while strictly preserving the source image's identity and appearance~\cite{ma2017poseguided,zhang2022dptn}. This task is important for commercial applications such as virtual try-on, digital avatars, and film production~\cite{zhu2024appearance}. Extensive studies have explored this field ~\cite{shen2024pcdms, lu2024cfld,liu2025mcld}, which typically use VAE~\cite{kingma2014autoencoding} and CLIP~\cite{radford2021clip} to extract visual details and semantic cues from the source image, and inject them into the generative network with the target pose. The network then denoises the latent representation to produce a target image with an appearance consistent with the source image, following the latent-diffusion generation paradigm~\cite{rombach2022latent}. However, when the target pose differs substantially from the source pose, the lack of source information makes it challenging to produce plausible results.


\begin{figure}[!t]
 \centering
    \includegraphics[width=1\columnwidth]{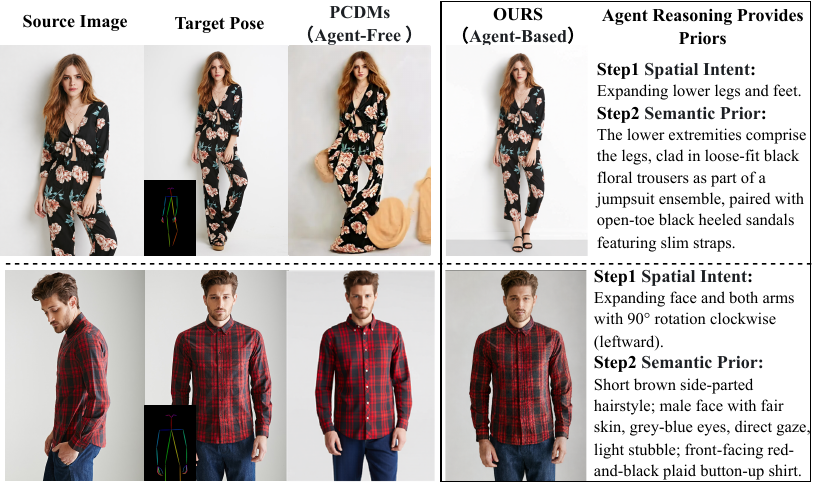}
\caption{Comparisons across two challenging cases.}
  \label{fig:fig1}
\end{figure}
Existing solutions to this problem can be broadly categorized into two distinct paradigms. (1) Priors from LLM. They leverage large language models to provide structural or textual guidance. For instance, PoseLLaVA~\cite{posellava} integrates SMPL-based pose representations into a multimodal LLaVA framework to align textual instructions with fine-grained 3D human poses, enabling language-guided pose estimation, generation, and adjustment. (2) Mapping relationship reasoning. They focus on modeling the geometric transformations between the source and target views. PoCoLD~\cite{han2023pocold} incorporates 3D annotations to construct rigid pose constraints, combining them with source appearance features to enforce cross-view correspondences. And IMAGPose~\cite{shen2024imagpose} leverages multi-view prior information to analyze view correlations across multiple source images, seamlessly embedding texture details into the diffusion process. However, both paradigms fail under drastic viewpoint shifts. Pure vision-driven approaches lack the necessary priors to resolve extreme perspective changes, causing network optimization to degenerate into blind pixel-level hallucination. Meanwhile, recent LLM-based methods remain rudimentary, treating LLMs as static, one-off text conditioners rather than interactive cognitive engines capable of dynamic reasoning.

In this paper, we categorize drastic viewpoint shifts into two distinct scenarios: \textit{Cross-View Pose Inversion} and \textit{Out-of-View Body Completion}. Both scenarios introduce severe challenges regarding visual artifacts and semantic misalignment in the generated results. (1) \textit{Cross-View Pose Inversion.} Existing methods~\cite{bhunia2023pidm,han2023pocold,shen2024pcdms} generate images whose appearance heavily depends on a source image. When a large viewpoint discrepancy exists between the source and target pose conditions, the source image provides insufficient appearance priors for self-occluded or transformed regions, making it difficult to infer details like clothing patterns on the back or side. For instance, the bottom row of Figure~\ref{fig:fig1} shows that even the strong PCDMs~\cite{shen2024pcdms} loses distinct button details after a sharp left turn. (2) \textit{Out-of-View Body Completion.} In scenarios where the target pose exposes regions completely occluded in the source view, source appearance priors provide inadequate guidance for synthesis. While existing methods~\cite{zhang2022dptn,bhunia2023pidm,shen2024pcdms} map the source image to the target pose, they fail to reliably infer details missing from the source. For instance, the top row of Figure~\ref{fig:fig1} shows that PCDMs~\cite{shen2024pcdms} generates a person's lower body with anomalously sized, unrealistic black floral trousers.

To bridge this gap, we introduce DAC-Pose, an agent-driven multimodal framework that reformulates single-view human generation as a collaborative dual-agent system. Moving beyond conventional passive pipelines, DAC-Pose orchestrates two synergistic agents: a Prior Semantic Reasoning (PSR) agent and a Discrepancy-Aware Visual Encoding (DAVE) agent. The PSR agent acts as a cognitive engine, leveraging collaborative reasoning to deduce the attributes of unseen regions and mitigate semantic ambiguity. Concurrently, DAVE acts as a specialized visual perception agent to explicitly quantify and encode viewpoint-induced spatial misalignments, feeding robust spatial constraints back into the generative process. As illustrated in Figure~\ref{fig:fig1}, this autonomous synergy enables superior synthesis under drastic viewpoint shifts. In the first row, the agent's detailed textual descriptions of the legs allow the denoising process to successfully recover the intricate patterns of the black floral trousers. In the second row, despite a severe leftward pose rotation, the agent generates informative descriptions of the man's face and red plaid shirt, effectively guiding the model through a large pose inversion to yield clearer cross-view reconstructions.

\begin{itemize}
     \item We propose DAC-Pose, a novel agent-driven multimodal framework that utilizes a collaborative dual-agent system to tackle the  challenge of severe spatial misalignment under drastic viewpoint shifts.
    \item We introduce the PSR agent to serve as the cognitive engine that deduces the fine-grained semantic attributes of unseen regions and overcomes the semantic ambiguity of non-agentic pipelines.
    \item We design the DAVE agent to quantify and encode pose-induced spatial misalignment, which constraints the generative network toward high-fidelity reconstruction of discrepant regions.
    \item We conduct extensive experiments on the DeepFashion and Market-1501 benchmark against state-of-the-art methods to validate its effectiveness. Encouraging results under drastic viewpoint shifts further highlight the importance of our contributions.
\end{itemize}

The rest of this article is organized as follows. Section \ref{sec:related_work} briefly introduces related work. Section \ref{m} introduces and discusses our proposed DAC-Pose architecture. Section \ref{sec:experiments} presents experimental results and compares them with other methods. Section \ref{dc} summarizes the entire paper.

\section{Related Work}
\label{sec:related_work}

This section outlines the technological evolution of pose-guided human image generation. While early methods relied on Generative Adversarial Networks (GANs)~\cite{goodfellow2014generative} and explicit spatial transformations, the field has transitioned to probabilistic diffusion models. More importantly, to overcome the inherent cognitive bottlenecks of pure vision-driven models, current research trajectories are increasingly gravitating toward leveraging Large Language Models (LLMs) and autonomous AI agents to assist the synthesis process.

\subsection{Vision-Driven Pose-Guided Generation}

In the early stages of development, GANs~\cite{goodfellow2014generative}  established fundamental distribution modeling and synthesis capabilities for human image generation~\cite{SHI2022108351}, yielding significant improvements in overall plausibility and appearance diversity. For instance, Siarohin et al.~\cite{siarohin2018defgan}, Esser et al.~\cite{Esser_2018_CVPR} and Ren et al.~\cite{ren2020gfla} adopted explicit spatial alignment techniques~\cite{SHEN2022108451,LUO2025111626} to transfer pixels from source images to target poses. Meanwhile, Ma et al.~\cite{ma2017poseguided} and Han et al.~\cite{han2019clothflow} employed multi-stage generation pipelines~\cite{KHATUN2023109246} that first align body parts and subsequently refine blurred boundary information. Recently, latent diffusion models (LDMs)~\cite{GONG2026113644} have demonstrated superior synthesis performance at both qualitative and quantitative levels. In particular, frameworks based on latent diffusion models (LDMs) have substantially enhanced detail fidelity and generation stability. For example, Bhunia et al.~\cite{bhunia2023pidm} and Wang et al.~\cite{wang2024disco} significantly improved identity consistency and clothing texture fidelity by aligning appearance textures with target poses. Similarly, Shen et al.~\cite{shen2024pcdms} and Lu et al.~\cite{lu2024cfld} further improved the reconstruction accuracy of complex clothing textures through multi-stage denoising architectures. Despite their notable success under mild pose variations, these methods suffer from implausible artifacts and semantic misalignments when faced with drastic viewpoint shifts, primarily due to the scarcity of visible source priors. Despite their notable success under mild pose variations, these non-agentic, pure vision-driven methods inevitably produce implausible artifacts and semantic misalignments when faced with drastic viewpoint shifts. Fundamentally, constrained by the scarcity of visible source priors, these pipelines lack the cognitive reasoning required to logically deduce unseen regions, forcing the network into blind pixel-level hallucination.

\subsection{From LLM Assistance to Agentic Paradigms}

To mitigate the aforementioned scarcity of visual priors, increasing research efforts are focused on leveraging Large Language Models (LLMs)~\cite{brown2020language} to enhance the text-to-image generation process. These methods primarily utilize textual priors or prompts to improve generation quality, enabling LLMs to act as high-level semantic controllers or capable agents that assist in visual generation tasks. For example, Fan et al.~\cite{fan2025one} employed LLMs to generate character description prompts combined with facial embedding techniques to enhance character consistency. Lee et al.~\cite{2026PointT2I} used LLMs to generate human body keypoints from text, subsequently utilizing diffusion models to produce high-fidelity human images that align with both the textual description and the specific pose. Feng et al.~\cite{posellava} drove pose estimation and optimization through detailed linguistic instructions, achieving precise alignment across pose, text, and visual modalities. However, such methods are typically static, offering only one-off prompt enhancements; they notably lack mechanisms for interactive reasoning and iterative optimization, thereby failing to explicitly construct spatial structure mappings or achieve effective cross-modal collaborative reasoning. 

Therefore, this paper proposes an Agent-based dual-system framework to address the aforementioned problems, aiming to utilize textual, pose, and visual information as complementary conditions. This enables the framework to break through the traditional LLMs-assisted generation paradigm, achieving interactive reasoning and cross-modal iterative optimization processes, thus improving the quality of structure-aware generation and multimodal consistency.



\begin{figure*}[!t]
 \centering
    \includegraphics[width=1\linewidth]{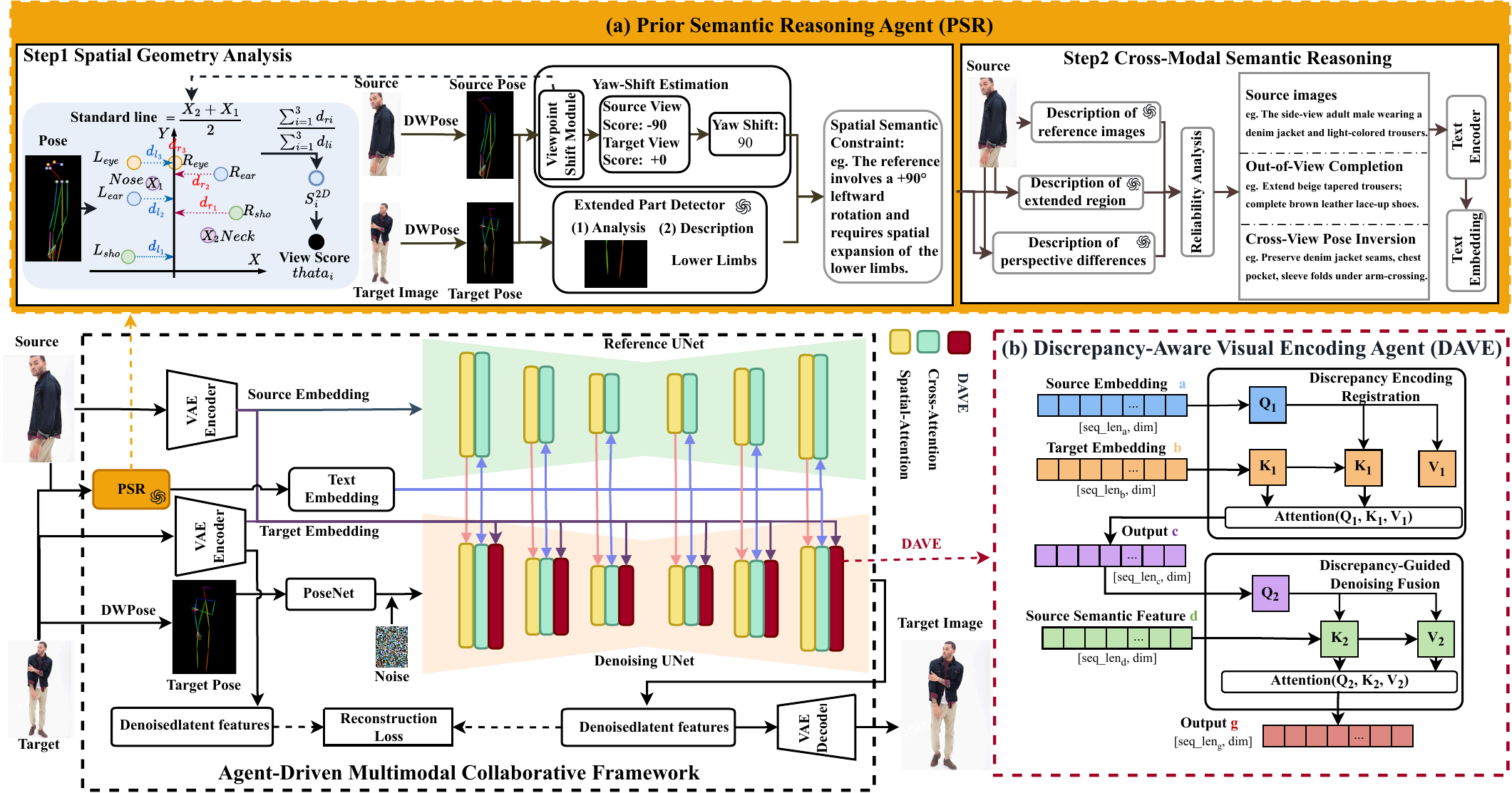}
    \caption{The overall architecture of DAC-Pose framework with (a) Prior Semantic Reasoning Agent and (b) Discrepancy-Aware Visual Encoding Agent.}
  \label{fig:fig2}
\end{figure*}

\section{Methodology}
\label{m}
\subsection{Overview}
This section introduces DAC-Pose, an agent-driven multimodal framework designed to resolve generation quality degradation caused by severe spatial misalignments between the source person and target pose. Existing non-agentic approaches generally fall into two paradigms. The first focuses on optimizing generation by strengthening the pose mapping between the source and target configurations~\cite{lu2024cfld,hu2024animateanyone,Xu_2024_CVPR}. However, bound by single-view limitations, these purely vision-driven methods lack the agentic cognition required to deduce unseen regions, failing when faced with large pose discrepancies~\cite{lu2024cfld,shen2024imagpose}. The second paradigm introduces multi-view images to comprehensively extract appearance priors and passively compensate for occlusions~\cite{shen2024imagpose}. Nevertheless, collecting multi-view data of a single identity is impractical, limiting its real-world applicability.

To overcome these limitations, we introduce DAC-Pose, a novel framework that reformulates single-view high-fidelity synthesis as a collaborative, dual-agent system. Within this paradigm, visual cues preserve identity and texture, pose signals dictate spatial layouts, and textual descriptions act as stable, explicitly reasoned semantic anchors~\cite{Liu_2023_NeurIPS,lu2024cfld}. Under single-view constraints, DAC-Pose is driven by two specialized autonomous agents: the \textbf{Prior Semantic Reasoning} (PSR) agent, a cognitive engine that logically deduces missing appearance attributes to construct robust semantic priors, and the \textbf{Discrepancy-Aware Visual Encoding} (DAVE) agent, a perception component that explicitly quantifies and encodes viewpoint-induced spatial misalignments. Fig.~\ref{fig:fig2} illustrates the overall architecture.

\subsection{Prior Semantic Reasoning Agent}

Existing non-agentic methods for human image synthesis predominantly rely on directly mapping the visible appearance of a source image onto a target pose condition~\cite{bhunia2023pidm,lu2024cfld,shen2024imagpose}. However, when faced with substantial pose discrepancies, a single-view source inherently fails to provide sufficient visual priors for view-shifted regions or newly exposed body parts~\cite{bhunia2023pidm,lu2024cfld}. Lacking agentic cognition and reliable structural constraints, these purely vision-driven generative networks are highly prone to semantic misalignments, localized artifacts, and implausible structural hallucinations~\cite{Ju_2023_ICCV,lu2024cfld}.

To overcome these cognitive deficits, we introduce the PSR agent to serve as the framework's cognitive engine, deducing explicit semantic priors for highly discrepant regions. Unlike direct pixel-level visual completion, textual descriptions can explicitly articulate fine-grained attributes, such as color, material, and inter-regional relationships~\cite{Liu_2023_NeurIPS,lu2024cfld}. This explicit text-driven formulation effectively mitigates texture instability and structural deviations, providing robust conditioning for the diffusion model. Furthermore, accurate semantic compensation for unseen regions hinges not only on the source appearance but also on the topological transformations between the source and target poses~\cite{bhunia2023pidm,Ju_2023_ICCV}.

To overcome the limitations of global text generation in capturing complex spatial dynamics, the PSR agent employs a collaborative reasoning mechanism. The agent first conducts \textit{Spatial Geometry Analysis} to quantify pose-induced spatial misalignments. It then performs \textit{Cross-Modal Semantic Inference} to logically fuse the source appearance with these spatial constraints. Specifically, the geometric analysis localizes structural discrepancies, while the semantic inference infers region-specific priors to compensate for missing appearance details. This synergistic design yields precise targeted textual priors, establishing robust semantic guidance for the downstream generative network.

\subsubsection{Spatial Geometry Analysis}

It transforms the pose discrepancy between the source and target pose conditions into explicit spatial directives. As shown in Fig.~\ref{fig:fig2}(a), we employ DWPose~\cite{yang2023dwpose} to extract 2D human skeletons from the source image $I_s$ and the target image $I_t$:
\begin{equation}
P_s=\mathcal{E}_{pose}(I_s), \quad P_t=\mathcal{E}_{pose}(I_t),
\end{equation}
where $\mathcal{E}_{pose}(\cdot)$ denotes the 2D pose estimator, and $P_s, P_t$ contain the human keypoint coordinates and visibility states. $P_s$ and $P_t$ denote the source pose map and the target pose map, respectively. Utilizing $P_s$ and $P_t$, \textit{Yaw-Shift Estimation} determines the viewpoint discrepancy, while the \textit{Extended Part Detector} localizes newly exposed or expanded body regions. Finally, \textit{Spatial Semantic Constraint} integrates these findings to construct the definitive spatial semantic constraints for downstream cross-modal semantic reasoning.

\textit{(a) Yaw-Shift Estimation.}
It estimates the yaw shift between the source pose and the target pose. It contains two steps. First, the Viewpoint Shift Module predicts the viewpoint score of each pose. Then, the source and target viewpoint scores are compared to obtain the yaw-shift result $D_{\mathrm{view}}$.

For pose $P_i$ ($i \in \{s, t\}$), the Viewpoint Shift Module constructs a 2D Cartesian coordinate system on the pose map. The nose and neck keypoints act as anchors defining the vertical reference axis. Given their horizontal coordinates $x_i^{\text{nose}}$ and $x_i^{\text{neck}}$, the horizontal position of the reference axis is formulated as:
\begin{equation}
x_i^{\mathrm{std}}=\frac{x_i^{\mathrm{nose}}+x_i^{\mathrm{neck}}}{2}.
\end{equation}

This reference axis measures the left-right distribution of symmetric keypoints, such as eyes, ears, and shoulders, which are defined as $\mathcal{K} = \{\text{eyes}, \text{ears}, \text{shoulders}\}$. For each pair $m \in \mathcal{K}$, let $x_{i,L_m}$ and $x_{i,R_m}$ denote the horizontal coordinates of the left and right keypoints, respectively. For example, if $m$ represents the eyes and $i=s$, then $x_{s,L_m}$ and $x_{s,R_m}$ correspond to the horizontal coordinates of the left and right eyes in the source image $I_s$. Their respective distances to the reference axis are defined as:
\begin{equation}
d_{i,L_m}=|x_i^{\mathrm{std}}-x_{i,L_m}|, \qquad
d_{i,R_m}=|x_{i,R_m}-x_i^{\mathrm{std}}|.
\end{equation}

Let $\mathcal{K}_i \subseteq \mathcal{K}$ be the set of valid pairs for $P_i$. The raw 2D viewpoint response is calculated as:
\begin{equation}
R_i^{2D}
=\frac{\sum_{m\in\mathcal{K}_i}\omega_m d_{i,rm}}
{\sum_{m\in\mathcal{K}_i}\omega_m d_{i,lm}+\varepsilon},
\end{equation}
where $\omega_m$ is the weight of pair $m$, satisfying $\sum_{m\in\mathcal{K}_i}\omega_m=1$, and $\varepsilon$ is a small constant introduced for numerical stability.

The raw response is then normalized to yield a bounded viewpoint score:
\begin{equation}
\hat{S}_i^{2D}
=\operatorname{sat}
\left(
\frac{R_i^{2D}-1}{R_i^{2D}+1+\varepsilon}
\right),
\end{equation}
where $\operatorname{sat}(\cdot)$ is a saturation function limiting the input to $[-1,1]$. Positive and negative scores indicate right- and left-side dominance, respectively, while a score close to zero indicates a nearly frontal pose.

The bounded score is then discretized into a distinct viewpoint state:
\begin{equation}
S_i=\mathcal{Q}_{view}(\hat{S}_i^{2D}),
\end{equation}
where $\mathcal{Q}_{view}(\cdot)$ denotes the viewpoint quantization function, and $S_i$ represents the viewpoint state predicted by the Viewpoint Shift Module for pose $P_i$.

By applying the Viewpoint Shift Module to both the source pose $P_s$ and target pose $P_t$, we obtain their corresponding viewpoint states, $S_s$ and $S_t$. The yaw shift is then calculated as:
\begin{equation}
D_{\mathrm{view}}=S_t-S_s,
\end{equation}
here, $D_{\mathrm{view}}$ records the direction and magnitude of the viewpoint transition from the source to the target pose, serving as the viewpoint-shift component in the subsequent \textit{Spatial Semantic Constraint}.

\textit{(b) Extended Part Detector.}
Its purpose is to detect body regions that emerge or spatially expand during the transition to the target pose. It consists of two steps: \textit{Analysis} and \textit{Description}.

In the \textit{Analysis} step, the LLM-based detector compares the source pose $P_s$ and target pose $P_t$ regarding their skeletal structures, body-part visibility, and local spatial extension. It then localizes the specific regions showing distinct expansion or new skeletal exposure in the target pose.

In the \textit{Description} step, the LLM-based detector converts the localized skeletal region into a concise semantic expression. The resulting output, denoted as $D_{\mathrm{exp}}$, records this expanded region in text form. For example, as shown in Fig.~\ref{fig:fig2}(a), the detected region is described as ``Lower Limbs''.

\textit{(c) Spatial Semantic Constraint.}
This step aims to organize the results of the Yaw-Shift Estimation and the Extended Part Detector into a unified spatial semantic constraint for subsequent semantic reasoning. This integration is implemented via an LLM-based process that combines the viewpoint-shift result $D_{\mathrm{view}}$ and the expanded-region description $D_{\mathrm{exp}}$ into:
\begin{equation}
C_{\mathrm{spa}}=\mathcal{G}_{\mathrm{spa}}(D_{\mathrm{view}},D_{\mathrm{exp}}),
\end{equation}
where $\mathcal{G}_{\mathrm{spa}}(\cdot)$ represents the LLM-based spatial semantic constraint process.$C_{\mathrm{spa}}$ contains the yaw-shift direction and the expanded body-part description.

In Fig.~\ref{fig:fig2}(a), the spatial semantic constraint is formulated as: 'The reference involves a $+90^\circ$ rotation and requires spatial expansion of the lower limbs.' This constraint is then forwarded to the Cross-Modal Semantic Reasoning Agent to guide semantic inference across view-shifted and newly exposed regions.

\subsubsection{Cross-Modal Semantic Reasoning}
It generates reliable region-level textual priors by combining the source appearance with the spatial semantic constraint. Taking the source image $I_s$ and the constraint $C_{\mathrm{spa}}$ as inputs, it produces a semantic condition for the downstream diffusion generation process. As illustrated in Fig.~\ref{fig:fig2}(a), the module comprises three MLLM-based description branches calibrated through reliability analysis and text encoding.

First, the source-image description branch extracts detailed appearance semantics from both the source image and the spatial semantic constraint:
\begin{equation}
T_{\mathrm{source}}=\mathcal{A}_{\mathrm{source}}(I_s,C_{\mathrm{spa}}),
\end{equation}
where $\mathcal{A}_{\mathrm{source}}(\cdot)$ represents the MLLM-based source-image description process. The output, $T_{\mathrm{source}}$, details the visible appearance attributes of the source person, including clothing category, color, material, style, and region-related details guided by $C_{\mathrm{spa}}$. This text provides the source appearance prior for preserving identity and clothing consistency.

Second, the extended-region description branch focuses on the body regions specified by the spatial semantic constraint:
\begin{equation}
T_{\mathrm{ext}}=\mathcal{A}_{\mathrm{ext}}(I_s,C_{\mathrm{spa}}),
\end{equation}
where $\mathcal{A}_{\mathrm{ext}}(\cdot)$ denotes the MLLM-based extended-region description process. $T_{\mathrm{ext}}$ describes the semantic attributes of newly exposed or spatially expanded regions. For example, when $C_{\mathrm{spa}}$ indicates lower-limb expansion, this branch infers the corresponding clothing category, color, and material based on the visible source appearance.

Finally, the perspective-difference description branch analyzes the semantic changes caused by viewpoint variation:
\begin{equation}
T_{\mathrm{diff}}=\mathcal{A}_{\mathrm{diff}}(I_s,C_{\mathrm{spa}}),
\end{equation}
where $\mathcal{A}_{\mathrm{diff}}(\cdot)$ denotes the MLLM-based perspective-difference description process, and $T_{\mathrm{diff}}$ characterizes the potential appearance of view-shifted regions while enforcing semantic consistency between the source appearance and the target-view structure.

After obtaining the three MLLM-generated descriptions, we perform a reliability analysis to evaluate their output confidence and semantic consistency:
\begin{equation}
T_{\mathrm{sem}}
=
\mathcal{F}_{\mathrm{rel}}(T_{\mathrm{source}}, T_{\mathrm{ext}}, T_{\mathrm{diff}}),
\end{equation}
where $\mathcal{F}_{\mathrm{rel}}(\cdot)$ denotes the reliability analysis function. The resulting $T_{\mathrm{sem}}$ aggregates the reference-image description, out-of-view body completion, and cross-view pose inversion cues, which are subsequently encoded by the text encoder.

The semantic compensation text is then mapped to a text embedding:
\begin{equation}
C_{\mathrm{sem}}=\Phi_{\mathrm{text}}(T_{\mathrm{sem}}),
\end{equation}
where $\Phi_{\mathrm{text}}(\cdot)$ denotes the text embedding encoder. $C_{\mathrm{sem}}$ is the resulting text embedding.

\subsection{Discrepancy-Aware Visual Encoding Agent}
Conventional non-agentic methods predominantly treat the target pose as a passive global condition, integrating it into the generative network without explicitly modeling cross-view spatial misalignments~\cite{Ju_2023_ICCV,lu2024cfld,hu2024animateanyone}. Consequently, in scenarios with severe pose variations, this passive conditioning hinders the network's ability to perceive localized spatial deformations. Lacking an active perception mechanism to track these shifts, such models frequently suffer from texture misalignment, blurred details, and anatomically implausible generation~\cite{Ju_2023_ICCV,lu2024cfld,Xu_2024_CVPR}.

To overcome this perceptual bottleneck, we introduce the DAVE agent as the specialized visual perception component of our dual-agent system. Instead of relying on passive feature extraction, DAVE explicitly quantifies and encodes viewpoint-induced spatial discrepancies between the source image and target pose. While extracting source appearance features, the agent actively constructs fine-grained discrepancy representations mapping the source person to target pose regions. These dynamic representations serve as auxiliary spatial constraints, continuously guiding the denoising network. By combining appearance priors with agent-driven pose-discrepancy relations, our framework improves awareness of localized topological transformations during synthesis.

\subsubsection{Discrepancy Encoding Registration}
It registers the spatial discrepancy between the source appearance and the target region to obtain an explicit representation of their regional correspondence during training.

We first extract the source appearance and target-region embeddings. Specifically, the source image $I_s$ and target image $I_t$ are encoded via a VAE encoder~\cite{rombach2022latent}:
\begin{equation}
a=\mathcal{E}_{\mathrm{vae}}(I_s), \quad b=\mathcal{E}_{\mathrm{vae}}(I_t),
\end{equation}
where $\mathcal{E}_{\mathrm{vae}}(\cdot)$ denotes the VAE encoder, while $a$ and $b$ represent the source appearance embedding and the target-region embedding used for discrepancy registration, respectively.

We then employ a cross-attention mechanism to model the correspondence between $a$ and $b$. Specifically, the source appearance embedding $a$ serves as the query, while the target-region embedding $b$ provides the key and value states:
\begin{equation}
c=\operatorname{CA}(a,b,b),
\end{equation}
where $c$ denotes the registered discrepancy encoding. By modeling the regional relationship between the source appearance and the target region, it serves as a crucial spatial discrepancy prior for the subsequent denoising phase.

\subsubsection{Discrepancy-Guided Denoising Fusion}

This module aims to fuse the registered discrepancy encoding with the source semantic feature, thereby forming a discrepancy-guided condition for the subsequent denoising process. After the first-stage training, the optimized reference branch provides the source semantic feature $d$ for the source image under the guidance of $C_{\mathrm{sem}}$.

The registered discrepancy encoding $c$ obtained from the previous module is then fused with the source feature through cross-attention. Specifically, $c$ is used as the query, while $d$ is used as the key and value:
\begin{equation}
g=\operatorname{CA}(c,d,d),
\end{equation}
where $g$ denotes the discrepancy-guided feature. This fusion incorporates the registered source-target discrepancy into the source feature representation.

Finally, the discrepancy-guided feature is used as the output condition of the DAVE block:
\begin{equation}
C_{\mathrm{DAVE}}=g.
\end{equation}
$C_{\mathrm{DAVE}}$ is used in the second-stage generation as a discrepancy-aware condition, enabling the denoising network to emphasize regions with large source-target discrepancies, such as viewpoint shifts and expanded body parts.

\section{Experiments}
\label{sec:experiments}

\subsection{Experimental Settings}
\label{sec:implementation}

\textbf{Datasets.}
Experiments are conducted on DeepFashion~\cite{liu2016deepfashion} and Market-1501~\cite{zheng2015market1501}. DeepFashion contains 52,712 high-resolution images of fashion models, which we resize to $256 \times 176$. Market-1501 consists of 32,668 low-resolution pedestrian images featuring significant variations in viewpoint, background, and illumination, which we resize to $128 \times 64$. For both datasets, poses are extracted using DWPose~\cite{yang2023dwpose}.

\textbf{Implementations.}
Experiments were conducted on eight 80 GB NVIDIA A100 GPUs. We adopt Stable Diffusion~\cite{rombach2022latent} as the diffusion backbone and GPT-5.5 as the multimodal foundation model for the PSR and DAVE agents. Sentence-BERT~\cite{reimers2019sentencebert} encodes the textual semantic priors from the PSR agent, and the lightweight Pose Guider from Animate Anyone~\cite{hu2024animateanyone} serves as our PoseNet. During training, the VAE and Sentence-BERT are frozen. Training spans 30,000 steps with a batch size of 64 using Adam~\cite{kingma2015adam} ($\text{LR} = 1 \times 10^{-5}$). For inference, DDIM~\cite{song2021ddim} executes 20 denoising steps.

\subsection{Comparison With State-of-the-Art Methods}
\label{sec:sota_comparison}

We evaluate DAC-Pose under two challenging scenarios featuring large discrepancies between the source image and target pose: (1) drastic viewpoint shift, which tests appearance preservation under severe viewpoint changes, and (2) out-of-view body completion, which tests structural plausibility when the target pose reveals body regions absent from the source. Primary experiments are conducted on DeepFashion, with cross-dataset generalization assessed on Market-1501~\cite{zheng2015market1501}. Evaluation metrics include SSIM~\cite{wang2004ssim} for structural preservation, alongside LPIPS~\cite{zhang2018lpips} and FID~\cite{heusel2017fid} for perceptual similarity and distributional realism, where higher SSIM and lower LPIPS/FID indicate superior performance. We compare DAC-Pose with Def-GAN~\cite{siarohin2018defgan}, PATN~\cite{zhu2019patn}, ADGAN~\cite{men2020adgan}, PISE~\cite{zhang2021pise}, GFLA~\cite{ren2020gfla}, DPTN~\cite{zhang2022dptn}, NTED~\cite{ren2022nted}, CASD~\cite{zhou2022cross}, PoCoLD~\cite{han2023pocold}, PIDM~\cite{bhunia2023pidm}, CFLD~\cite{lu2024cfld}, PCDMs~\cite{shen2024pcdms}, PMMD~\cite{shang2026pmmd}, and IMAGPose~\cite{shen2024imagpose} under unified experimental settings.

In our evaluation, rather than comparing against rudimentary text-prompting heuristics, we benchmark DAC-Pose against state-of-the-art (SOTA) pure vision-driven and multi-view enhanced frameworks (e.g., PCDMs). This rigorous comparison isolates and validates our core hypothesis: that explicit agentic reasoning is fundamentally superior to passive feature alignment for resolving drastic viewpoint shifts, establishing a clear contrast between our cognitive system and existing non-agentic spatial mapping methods.

\begin{table*}[!htbp]
  \centering
  \normalsize
  \setlength{\tabcolsep}{8pt}
  \begin{tabular}{llccc}
    \toprule
    Dataset & Method & SSIM $\uparrow$ & LPIPS $\downarrow$ & FID $\downarrow$ \\
    \midrule
    \multirow{15}{*}{\shortstack{DeepFashion\\($256 \times 176$)}}
      & Def-GAN~\cite{siarohin2018defgan} & 0.6786 & 0.2330 & 18.457 \\
      & PATN~\cite{zhu2019patn}           & 0.6709 & 0.2562 & 20.751 \\
      & ADGAN~\cite{men2020adgan}         & 0.6721 & 0.2283 & 14.458 \\
      & PISE~\cite{zhang2021pise}         & 0.6629 & 0.2059 & 13.610 \\
      & GFLA~\cite{ren2020gfla}           & 0.7074 & 0.2341 & 10.573 \\
      & DPTN~\cite{zhang2022dptn}         & 0.7112 & 0.1931 & 11.387 \\
      & NTED~\cite{ren2022nted}           & 0.7182 & 0.1752 & 8.6838 \\
      & CASD~\cite{zhou2022cross}          & 0.7248 & 0.1936 & 11.373 \\
      & PoCoLD~\cite{han2023pocold}       & 0.7310 & 0.1642 & 8.0667 \\
      & PIDM~\cite{bhunia2023pidm}        & 0.7312 & 0.1678 & 6.3671 \\
      & CFLD~\cite{lu2024cfld}            & 0.7378 & 0.1519 & 6.8040 \\
      & PCDMs~\cite{shen2024pcdms}        & 0.7444 & 0.1365 & 7.4734 \\
      & PMMD~\cite{shang2026pmmd}         & 0.7492 & 0.1322 & 6.9475 \\
      & IMAGPose~\cite{shen2024imagpose}  & 0.7561 & 0.1284 & 5.8738 \\
      & \textbf{DAC-Pose (Ours)}          & \textbf{0.7572} & \textbf{0.1274} & \textbf{5.8547} \\
    \midrule
    \multirow{7}{*}{\shortstack{Market-1501\\($128 \times 64$)}}
      & Def-GAN~\cite{siarohin2018defgan} & 0.2683 & 0.2994 & 25.364 \\
      & PATN~\cite{zhu2019patn}           & 0.2821 & 0.3196 & 22.657 \\
      & GFLA~\cite{ren2020gfla}           & 0.2883 & 0.2817 & 19.751 \\
      & DPTN~\cite{zhang2022dptn}         & 0.2854 & 0.2711 & 18.995 \\
      & PIDM~\cite{bhunia2023pidm}        & 0.3054 & 0.2415 & 14.451 \\
      & PCDMs~\cite{shen2024pcdms}        & 0.3169 & 0.2238 & 13.897 \\
      & \textbf{DAC-Pose (Ours)}          & \textbf{0.3282} & \textbf{0.2104} & \textbf{12.659} \\
    \bottomrule
  \end{tabular}
  \caption{Quantitative comparison on DeepFashion and Market-1501 at different image resolutions.}
  \label{tab:quantitative_comparison}
\end{table*}

\subsubsection{Quantitative Comparison}

As shown in Table~\ref{tab:quantitative_comparison}, DAC-Pose achieves the best performance across all three metrics on DeepFashion, yielding an SSIM of 0.7572, an LPIPS of 0.1274, and an FID of 5.8547. Compared with the strongest competing framework, IMAGPose, DAC-Pose delivers steady gains: it advances SSIM from 0.7561 to 0.7572, and lowers LPIPS and FID from 0.1284 and 5.8738 to 0.1274 and 5.8547, respectively. The performance gains become even more pronounced when compared against PCDMs. Specifically, DAC-Pose improves SSIM by 0.0128 and reduces LPIPS and FID by 0.0091 and 1.6187, respectively. These results demonstrate that DAC-Pose preserves the target structure more effectively than PCDMs while maintaining high appearance fidelity.

\begin{figure*}[t]
  \centering
  \includegraphics[width=\textwidth]{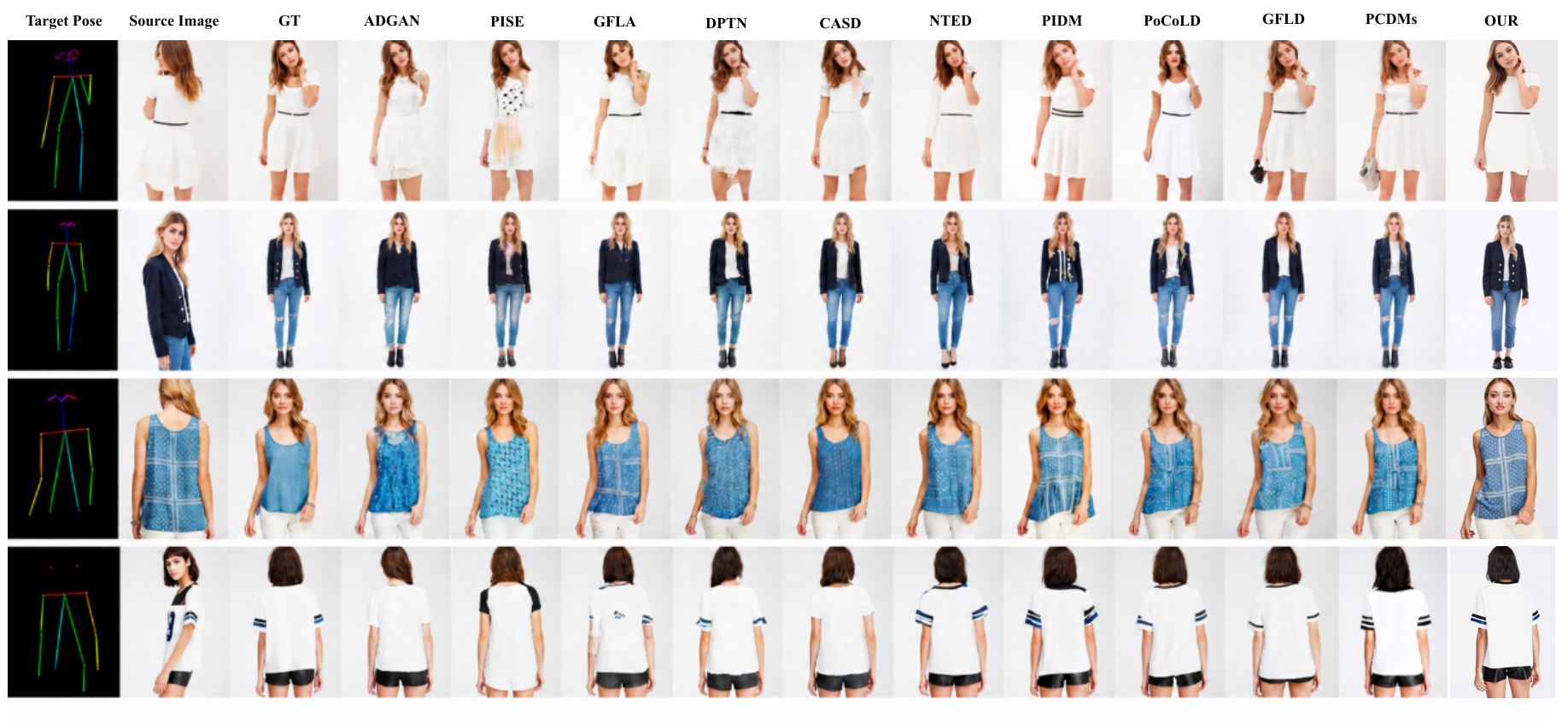}
  \caption{Qualitative comparison under cross-view pose inversion on DeepFashion.}
  \label{fig:viewpoint_shifts}
\end{figure*}
\begin{figure*}[t]
  \centering
  \includegraphics[width=\textwidth]{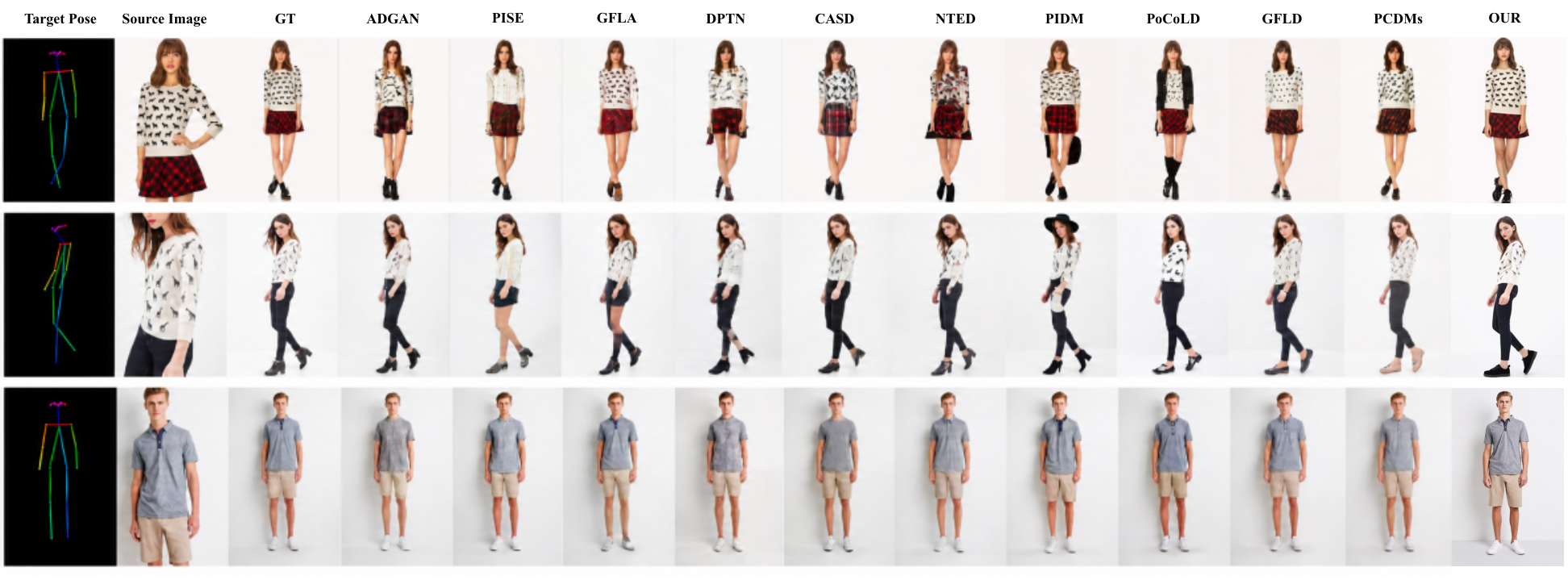}
  \caption{Qualitative comparison under out-of-view body completion on DeepFashion.}
  \label{fig:out_of_view_completion}
\end{figure*}

Table~\ref{tab:quantitative_comparison} also presents evaluation results on Market-1501, where DAC-Pose outperforms all compared methods with an SSIM of 0.3282, LPIPS of 0.2104, and FID of 12.659. Compared to the strongest baseline PCDMs, DAC-Pose advances SSIM by 0.0113 and lowers LPIPS and FID by 0.0134 and 1.238, respectively. The steady gains across both DeepFashion and Market-1501 validate the robustness of our framework when generalizing to datasets with varying resolutions, qualities, and capture environments.

\subsubsection{Qualitative Comparison}

Figures~\ref{fig:viewpoint_shifts} and~\ref{fig:out_of_view_completion} present qualitative results for cross-view pose inversion and out-of-view body completion, respectively.

\textbf{Cross-View Pose Inversion.}
Figure~\ref{fig:viewpoint_shifts} illustrates model performance under severe viewpoint discrepancies. In the first row, where the target requires synthesizing a front view from a back-view source image, ADGAN produces a blurred dress silhouette with an unclear waistline, while PISE introduces dark spots and yellowish artifacts. In contrast, DAC-Pose reconstructs the garment boundaries and silhouette with superior visual consistency. In the second row, PISE and NTED distort the jacket boundary and trouser details, whereas PIDM introduces artifacts like overly prominent front buttons. DAC-Pose preserves a coherent texture arrangement within the newly visible regions. The fourth row evaluates side-to-back synthesis of a patterned T-shirt, where GFLA and PIDM incorrectly transfer front-view visual cues to the generated back view. DAC-Pose generates a clean back-view appearance, maintaining consistent sleeve stripes and garment boundaries. Collectively, these observations demonstrate that DAC-Pose effectively suppresses texture leakage, color drift, and boundary distortion during cross-view pose inversion.

\textbf{Out-of-View Body Completion.}
Figure~\ref{fig:out_of_view_completion} presents three examples where the source image captures only a partial body region, while the target pose requires a complete human image. In the first row, CASD and NTED distort the black pattern on the white sweater, whereas DPTN introduces an unprompted handbag absent from the source. In the second row, PISE alters the dark trousers into shorts, PIDM introduces extraneous accessories (a hat and handbag), and PCDMs modify the appearance of the lower garment. In contrast, DAC-Pose synthesizes plausible completions of the limbs and lower garments while faithfully preserving the clothing category, color distribution, and identity-specific appearance of the source image.

\subsection{Ablation Study}
\label{sec:ablation_study}

To validate each component, we perform ablation studies on DeepFashion by evaluating two variants: w/o PSR (omitting prior semantic reasoning) and w/o DAVE (omitting discrepancy-aware visual encoding).

\begin{table}[!t]
  \centering
  \normalsize
  \begin{tabular}{lccc}
    \toprule
    Methods & SSIM $\uparrow$ & LPIPS $\downarrow$ & FID $\downarrow$ \\
    \midrule
    w/o DAVE & 0.7306 & 0.1442 & 6.5116 \\
    w/o PSR  & 0.7428 & 0.1315 & 6.2380 \\
    \textbf{DAC-Pose (Ours)}
      & \textbf{0.7572}
      & \textbf{0.1274}
      & \textbf{5.8547} \\
    \bottomrule
  \end{tabular}
  \caption{Quantitative ablation on DeepFashion.}
  \label{tab:ablation}
\end{table}

\subsubsection{Quantitative Ablation Analysis}

As detailed in Table~\ref{tab:ablation}, removing the PSR agent yields a clear performance drop, decreasing SSIM from 0.7572 to 0.7428, while increasing LPIPS and FID to 0.1315 and 6.2380, respectively. This decline confirms that the cognitive semantic priors generated by PSR are crucial for guiding the synthesis process. Notably, eliminating the DAVE agent incurs an even steeper performance drop, deteriorating SSIM to 0.7306, LPIPS to 0.1442, and FID to 6.5116. This pronounced drop highlights that pure semantic reasoning cannot guarantee visual fidelity alone, proving that DAVE's discrepancy-aware constraints are indispensable for structural alignment and visual consistency under severe pose changes.

\begin{figure}[!t]
  \centering
  \includegraphics[width=\columnwidth]{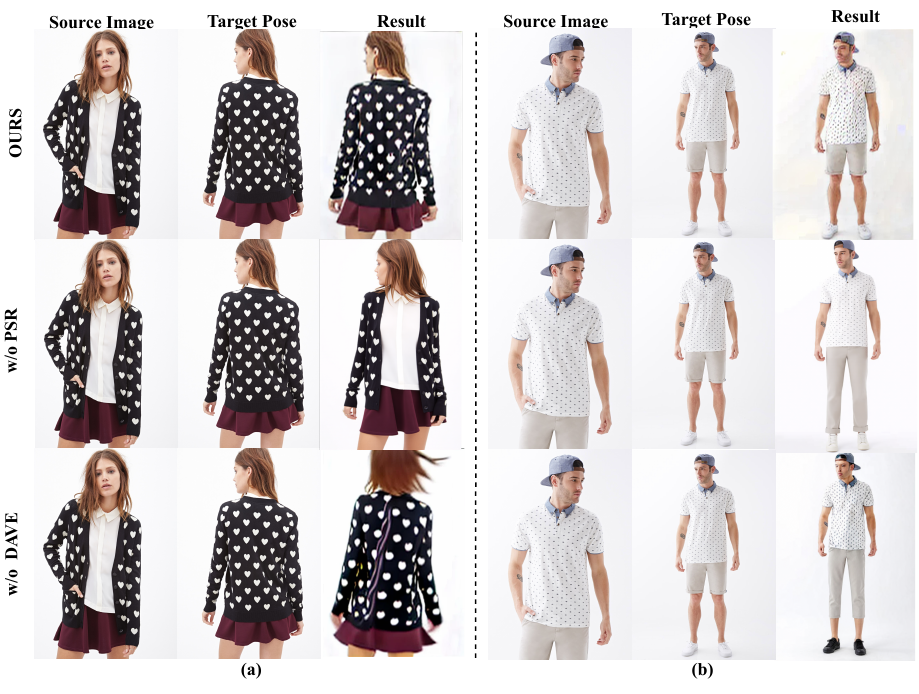}
  \caption{Qualitative ablation under (a) cross-view pose inversion and (b) out-of-view body completion.}
  \label{fig:qualitative_ablation}
\end{figure}

\subsubsection{Qualitative Ablation Analysis}

Figure~\ref{fig:qualitative_ablation} presents a visual comparison of the full DAC-Pose framework against its variants lacking either the PSR or DAVE agent. To evaluate the synergistic effects of these components, we analyze their generative performance across two challenging scenarios: cross-view pose inversion and out-of-view body completion.

\textbf{Cross-View Pose Inversion.} 
As depicted in figure~\ref{fig:qualitative_ablation}(a), the target pose demands synthesizing a back view from a strictly frontal source image. The full model successfully preserves the delicate white heart-shaped patterns on the black jacket while generating a structurally coherent back-view topology. However, removing the PSR agent causes the generated images to retain front-view clothing attributes, such as the open jacket and white inner shirt. This indicates that, without PSR, the model fails to infer the appearance changes induced by viewpoint reversal, leading to severe semantic inconsistency. Conversely, removing the DAVE agent preserves the back-view semantics but yields severely distorted heart-shaped patterns, demonstrating the model's inability to maintain spatial fidelity without active visual perception. These observations demonstrate that PSR and DAVE provide complementary and indispensable constraints for handling cross-view pose inversion.

\textbf{Out-of-View Body Completion.} 
Figure~\ref{fig:qualitative_ablation}(b) illustrates a scenario where the target pose exposes a substantially larger body region than the truncated source image. The full DAC-Pose framework accurately preserves the intricate patterns on the white shirt, the color of the shorts, and the overall proportions. In contrast, removing the PSR agent causes the framework to generate long pants instead of shorts, indicating a clear deficiency in logical semantic completion. Meanwhile, omitting the DAVE agent introduces incorrect details, such as ankle-length pants and black shoes, reflecting a decline in visual fidelity and fine-detail generation. These results demonstrate that the PSR agent is crucial for accurate semantic reasoning, while the DAVE agent is essential for preserving appearance details during out-of-view body completion.

\section{Conclusion}
\label{dc}
In this paper, we presented DAC-Pose, an agent-driven multimodal framework for single-view pose-guided human image generation, with a focus on challenges involving cross-view pose inversion and out-of-view body completion. DAC-Pose introduces a Prior Semantic Reasoning (PSR) agent to infer semantic priors for regions that are absent in the source image. It incorporates a Discrepancy-Aware Visual Encoding (DAVE) agent to model source-target pose discrepancies and provide spatial constraints for the generative process. By integrating semantic reasoning with discrepancy-aware visual encoding, DAC-Pose improves the synthesis of plausible human appearance under drastic viewpoint shifts. Experiments on DeepFashion and Market-1501 show that DAC-Pose achieves competitive improved performance over existing methods in terms of SSIM, LPIPS, and FID. Qualitative comparisons further demonstrate its advantages in cross-view pose inversion and out-of-view body completion. Ablation studies verify the complementary contributions of PSR and DAVE, showing that both semantic priors and discrepancy-aware constraints are important for the proposed framework.

\section{Acknowledgments}
This work was supported by the Science and Technology Development Fund, Macao SAR, under the Basic Research Program
(0006/2024/RIA1), Science and Technology Development Fund--Ministry of Science and Technology under the National Key R\&D Program of China (0007/2025/AMJ and 2025YFE0202900), Shenzhen Science Fund for Excellent Young Scholars (RCYX20221008093036022), Special Support Plan for Outstanding Young Talents of Guangdong Province (2023TQ
07L745), and the Youth Innovation Promotion Association of the Chinese Academy of Sciences (2021358).
\bibliographystyle{elsarticle-num}
\bibliography{ref}

\end{document}